# Accurate 3D Reconstruction of Dynamic Scenes from Monocular Image Sequences with Severe Occlusions[*]

Vladislav Golyanik    Torben Fetzer    Didier Stricker

Department of Computer Science, University of Kaiserslautern
Department Augmented Vision, DFKI

{Vladislav.Golyanik, Torben.Fetzer, Didier.Stricker}@dfki.de

## Abstract

*The paper introduces an accurate solution to dense orthographic Non-Rigid Structure from Motion (NRSfM) in scenarios with severe occlusions or, likewise, inaccurate correspondences. We integrate a shape prior term into variational optimisation framework. It allows to penalize irregularities of the time-varying structure on the per-pixel level if correspondence quality indicator such as an occlusion tensor is available. We make a realistic assumption that several non-occluded views of the scene are sufficient to estimate an initial shape prior, though the entire observed scene may exhibit non-rigid deformations. Experiments on synthetic and real image data show that the proposed framework significantly outperforms state of the art methods for correspondence establishment in combination with the state of the art NRSfM methods. Together with the profound insights into optimisation methods, implementation details for heterogeneous platforms are provided.*

## 1. Introduction

Recovering a time varying geometry of non-rigid scenes from monocular image sequences is a fundamental, actively researched, yet a still unsolved problem in computer vision. Two main classes of approaches addressing it — template-based reconstruction and Non-Rigid Structure from Motion (NRSfM) — proved to be most effective so far. In the template-based reconstruction, scene geometry for at least one frame is known in advance, whereas in NRSfM no such information is given. Solely motion and deformation of a scene serve as reconstruction cues. Thereby, estimation of point correspondences in a pre-processing step is required. Measurement matrix combining correspondences is either obtained through a sparse keypoint tracking or a dense tracking of all visible points with optical flow.

NRSfM methods made significant advances during recent years in terms of the ability to reconstruct realistic non-rigid motion, especially for image sequences and motion capture data acquired in a controlled environment. Along with methods supporting an orthographic camera model [47, 29, 30, 7, 27, 15, 41, 31, 5], there are methods supporting a full perspective (in most of the cases calibrated) camera model [50, 21, 8, 23, 52, 49, 6, 12], dense reconstructions [36, 15, 2], sequential processing [26, 43, 1, 4, 3] and compound scenes [37]. At the same time, NRSfM is a highly ill-posed inverse problem in the sense of Hadamard, *i.e.*, the condition on the uniqueness of the solution is violated. In practice, a prior knowledge is required to disambiguate the solution space such as metric constraints [27], constraints on point trajectories [7, 53, 48, 4], temporal consistency assumption [15], local rigidity assumption [44, 31], soft inextensibility constraint [49, 12, 3], shape prior [14, 42, 41, 43] or the assumption on a compliance with a physical deformation model [4, 6].

Nevertheless, support for real-world image sequences is still limited due to the systematic violation of assumptions on the degree as well as the type of motion and deformation presented in a scene. Moreover, severe self- and external occlusions occur frequently, which results in noisy and erroneous correspondences. Since methods for computing correspondences are limited in compensating for occlusions, NRSfM methods should be able to cope with missing data and the associated disturbing effects robustly.

In this paper, a novel dense orthographic NRSfM approach is proposed which can cope with severe occlusions — Shape Prior based Variational Approach (SPVA) — along with a scheme for obtaining a shape prior from several non-occluded frames. The latter relies on a realistic assumption that a scene is not-occluded in a reference frame and there are some non-occluded views. Influence of the shape prior can be controlled by series of occlusion maps — an occlusion tensor — obtained from a measurement matrix and an input image sequence. In contrast to template-based reconstruction, the shape prior is computed automatically in our framework, and we do not rely on the rigidity assumption. The proposed methods are combined into a joint correspondence computation, occlusion detection, shape prior estimation and surface recovery *framework*, and evaluated against different state of the art non-rigid recovery pipeline configurations. SPVA surpasses state of the art in real scenarios with large occlusions or noisy correspondences, both in terms of the reconstruction accuracy and processing time. To the best of our knowledge, our method is the first to stably handle severe external occlusions in dense scenarios without requiring an expensive correspondence correction step.

---

[*]This work was supported by the project DYNAMICS (01IW15003) of the German Federal Ministry of Education and Research (BMBF).

## 2. Related Work

The proposed method is based on factorizing the measurement matrix into shapes and camera motion and operates on an image batch. The idea of factorisation was initially proposed for the rigid case [46] and adopted for the non-rigid case in [11] where every shape is represented by a linear combination of basis shapes. This statistical constraint can be interpreted as a basic form of a shape prior, and reflects the assumption on the linearity of deformations. This setting is known to perform well for moderate deformations and many successor methods built upon the idea of metric space constraints [47, 27, 36, 31]. In contrast, SPVA determines optimal basis shapes implicitly by penalizing nuclear norm of the shape matrix as proposed in [13].

For robustness to occlusions and missing data, several policies were proposed so far. One is to compensate for disturbing effects in the preprocessing step. Associating image points with their entire trajectories over an image sequence, Multi-Frame Optical Flow (MFOF) methods allow to detect occlusions and robustly estimate correspondences in occluded regions [16, 32, 33, 40]. These methods perform well if occlusions are rather small or of a short duration. Support of longer occlusions is, however, limited which results in reduced accuracy of NRSfM methods.

Another policy is to account for missing data and incorrect correspondences during surface recovery. In [47], Gaussian noise in measurements is explicitly modelled in the motion model. Authors report accurate results on perturbed inputs with an additive normally distributed noise. The shape manifold learning approach of [41] is withstandable against Gaussian noise (levels up to $12\%$ result in reconstructions of a decent accuracy). A method based on the recently introduced low-rank force prior includes a term accounting for a Gaussian noise in the measurements and was shown to handle $11.5\%$ of missing data caused by short-time occlusions [4]. Due to a variational formulation, the approach of Garg *et al.* [15] can compensate for small amount of erroneous correspondences, provided an appropriate solution initialisation is given. Due to a mode shape interpretation, the method of Agudo *et al.* [3] can perform accurately when $40\%$ of points are randomly removed from the input. Some other methods can also handle noisy and missing correspondences [27, 5], but in scenarios limited to short and local occlusions. In contrast, our method can cope with large and long occlusions.

Some NRSfM approaches allow integration of an explicit shape prior into the surface recovery procedure. Del Bue [14] proposed to jointly factorize measurement matrix and a pre-defined shape prior. The method showed enhanced performance under degenerate non-rigid deformations. The shape prior represented a single predefined static shape acquired by an external procedure or pre-computed basis shapes. Tao *et al.* [41] proposed to adopt a graph-based manifold learning technique based on diffusion maps where the shapes are constrained to lie on the pre-computed non-linear shape prior manifold. In this scheme, the basis shapes can be different for every frame and hence the method can reconstruct strong deformations. However, the approach requires a representative training set with a computationally expensive procedure (especially for the case of dense reconstructions) for embedding of new shapes not presented in the training set. Recent template-based reconstruction methods employ a similar principle as us [51, 22]. Thus, Yu *et al.* proposed to estimate a template shape at rest from several first frames provided sufficient cues for a multi-view reconstruction [51]. This estimate is based on rigidity assumption and the accuracy of the method depends on this step; an external pre-aligned template can also be used. Similarly, our approach estimates a shape prior from several initial frames. We also assume the initial views to be occlusion-free, but our method neither assumes rigidity nor requires a known template.

In our core approach, an estimated shape prior is integrated into a joint variational framework. It is most closely related to [13] due to the nuclear norm, and Variational Approach (VA) [15] due to the spatial integrity, *i.e.*, Total Variation (TV). Additionally, our energy functional includes a soft shape prior term. Camera poses are recovered in a closed-form through the projection of affine approximations on the SO(3) manifold (which is up to two orders of magnitude faster than non-linear optimisation). To detect occlusions, we propose a novel lightweight scheme relying on [16]. This approach differs from Taetz *et al.* [40] which corrects measurements in the pre-processing step but requires multiples of the computational time compared to [16].

## 3. Proposed Core Approach (SPVA)

Suppose $N$ points are tracked throughout an image sequence with $F$ frames and the input is assembled in the measurement matrix $\mathbf{W} \in \mathbb{R}^{2F \times N}$ so that every pair of rows contains $x$ and $y$ coordinates of a single frame respectively. During scene acquisition, an orthographic camera observes a non-rigidly deforming 3D scene $\mathbf{S} \in \mathbb{R}^{3F \times N}$.

Similarly, every treble of rows contains $x$, $y$ and $z$ coordinates of an instantaneous scene. $\mathbf{W}$ depends on a scene, relative camera poses as well as a camera model as

$$\mathbf{W} = \mathbf{P}\mathbf{R}_{3D}\mathbf{S}, \qquad (1)$$

where $\mathbf{R}_{3D} \in \mathbb{R}^{3F \times 3F}$ is a block-diagonal matrix with camera poses for every frame and $\mathbf{P} \in \mathbb{R}^{2F \times 3F}$ is a combined camera projection matrix with entries $\begin{pmatrix} 1 & 0 & 0 \\ 0 & 1 & 0 \end{pmatrix}$. Here, we additionally assume that the measurements are registered to the origin of the coordinate system and translation is resolved. The objective is to reconstruct a time varying shape $\mathbf{S}$ and relative camera poses $\mathbf{R}_{3D}$. In other words, we seek a realistic factorisation of $\mathbf{W}$. Since the third dimension is lost during the projection, $\mathbf{W}$ will be factorised in $\mathbf{P}\mathbf{R}_{3D} = \mathbf{R} \in \mathbb{R}^{2F \times 3F}$ and $\mathbf{S}$. In a post-processing step, $\mathbf{R}_{3D}$ can be estimated by imposing orthonormality constraints on rotation matrices, *i.e.*, entries of $\mathbf{R}$.

If additional information about shape of a scene is available, it can be used to constrain the solution space. We formulate NRSfM as a variational energy minimisation problem, and the most natural form of the shape prior is $\mathbf{S}_{\text{prior}} \in \mathbb{R}^{3F \times N}$, *i.e.*, a matrix containing prior shapes for every frame. In SPVA, $\mathbf{S}_{\text{prior}}$ influences the optimisation procedure in a flexible manner according to the required per frame and per pixel control. Next, depending on the control granularity level, several energies are proposed, and for each energy, an optimisation method is derived.

### 3.1. Per Sequence Shape Prior

A per-sequence shape prior is the strongest prior, *i.e.*, it allows to constrain the solution space for the whole sequence at once. Minimizer has the simplest form among all types, and the shape prior term has only a single weight parameter $\gamma$. The energy takes on the following form:

$$\underset{\mathbf{R},\mathbf{S}}{\operatorname{argmin}} \frac{\lambda}{2}\|\mathbf{W} - \mathbf{R}\mathbf{S}\|_{\mathcal{F}}^2 + \frac{\gamma}{2}\|\mathbf{S} - \mathbf{S}_{\text{prior}}\|_{\mathcal{F}}^2 + \\ + \sum_{f,i,p} \|\nabla \mathbf{S}_f^i(p)\| + \tau \|\mathrm{P}(\mathbf{S})\|_*, \qquad (2)$$

where $\sum_{f,i,p} \|\nabla \mathbf{S}_f^i(p)\|$ denotes TV with the gradient $\nabla \mathbf{S}_f^i(p)$ of the shape $\mathbf{S}_f$, $f \in \{1, \ldots, F\}$ at the point $p \in \{1, \ldots, N\}$ in the direction $i$; $\|\cdot\|_*$ and $\|\cdot\|_{\mathcal{F}}$ denote nuclear and Frobenius norms respectively, and the operator $\mathrm{P}(\cdot)$ permutes $\mathbf{S}$ into the matrix of the dimensions $F \times 3N$ (the point coordinates are rearranged framewise into single rows). The energy in Eq. (2) contains data, shape prior, smoothness and linear subspace model terms respectively.

If $\mathbf{R}$ or $\mathbf{S}$ is fixed, the energy is convex in $\mathbf{S}$ and $\mathbf{R}$ variables respectively. Such kind of energies, also called biconvex, can be optimised by Alternating Convex Search (ACS). In ACS, optimisation is performed for $\mathbf{R}$ and $\mathbf{S}$ while $\mathbf{S}$ or $\mathbf{R}$ is respectively fixed. Suppose $\mathbf{S}$ is fixed. In this case, the only term which depends on $\mathbf{R}$ is the data term. We seek a solution to the problem

$$\underset{\mathbf{R}}{\operatorname{argmin}} \frac{\lambda}{2}\|\mathbf{W} - \mathbf{R}\mathbf{S}\|_{\mathcal{F}}^2. \qquad (3)$$

The idea is to find an unconstrained solution $\mathbf{A}$ minimizing Eq. (3) and to project it blockwise into the SO(3) group in a closed-form. The projection will yield an optimal rotation matrix $\mathbf{R}$ [24]. First, we consider the sum of the separate data terms for every frame $f$ in the transposed form:

$$\sum_f \|\mathbf{W}_f^\mathsf{T} - \mathbf{S}_f^\mathsf{T}\mathbf{R}_f^\mathsf{T}\|_{\mathcal{F}}^2. \qquad (4)$$

Here, the property of invariance of Frobenius norm under transposition is used. Now an optimal matrix $\mathbf{A}_f$ can be found which minimizes the data term in Eq. (4) by projecting $\mathbf{W}_f^\mathsf{T}$ onto the column space of $\mathbf{S}_f^\mathsf{T}$ in a closed form:

$$\mathbf{A}_f = (\mathbf{S}_f\mathbf{S}_f^\mathsf{T})^{-1}\mathbf{S}_f^\mathsf{T}\mathbf{W}_f^\mathsf{T}. \qquad (5)$$

Note that the matrix $\mathbf{S}_f\mathbf{S}_f^\mathsf{T}$ has dimensions $3 \times 3$ which supports a low memory complexity of the optimisation. Next, we decompose $\mathbf{A}_f^\mathsf{T}$ with singular value decomposition (svd) and find $\mathbf{R}_f$ as follows:

$$\operatorname{svd}(\mathbf{A}_f^\mathsf{T}) = \operatorname{svd}(\mathbf{W}_f\mathbf{S}_f(\mathbf{S}_f\mathbf{S}_f^\mathsf{T})^{-1}) = \mathbf{U}\boldsymbol{\Sigma}\mathbf{V}^\mathsf{T} \qquad (6)$$

$$\mathbf{R}_f = \mathbf{U}\mathbf{C}\mathbf{V}^\mathsf{T}, \qquad (7)$$

where $\mathbf{C} = \operatorname{diag}(1, 1, \ldots, 1, \operatorname{sign}(\det(\mathbf{U}\mathbf{V}^\mathsf{T})))$. We favour the least squares solution for the sake of computational efficiency (see Sec. 5 for implementation details).

Next, we consider the energy functional in Eq. (2) with a fixed $\mathbf{R}$. We seek a solution to the problem

$$\underset{\mathbf{S}}{\operatorname{argmin}} \frac{\lambda}{2}\|\mathbf{W} - \mathbf{R}\mathbf{S}\|_{\mathcal{F}}^2 + \frac{\gamma}{2}\|\mathbf{S} - \mathbf{S}_{\text{prior}}\|_{\mathcal{F}}^2 + \\ \sum_{f,i,p} \|\nabla \mathbf{S}_f^i(p)\| + \tau \|\mathrm{P}(\mathbf{S})\|_*. \qquad (8)$$

This minimisation problem is convex, but it involves different norms and therefore cannot be solved in the standard way. After applying proximal splitting, we obtain two sub-problems with an auxiliary variable $\bar{\mathbf{S}}$:

$$\underset{\mathbf{S}}{\operatorname{argmin}} \frac{1}{2\theta}\|\mathbf{S} - \bar{\mathbf{S}}\|_{\mathcal{F}}^2 + \frac{\gamma}{2}\|\mathbf{S} - \mathbf{S}_{\text{prior}}\|_{\mathcal{F}}^2 + \\ \frac{\lambda}{2}\|\mathbf{W} - \mathbf{R}\mathbf{S}\|_{\mathcal{F}}^2 + \sum_{f,i,p}\|\nabla \mathbf{S}_f^i(p)\| \qquad (9)$$

$$\underset{\bar{\mathbf{S}}}{\operatorname{argmin}} \frac{1}{2\theta}\|\mathbf{S} - \bar{\mathbf{S}}\|_{\mathcal{F}}^2 + \tau\|\mathrm{P}(\bar{\mathbf{S}})\|_*. \qquad (10)$$

The minimisation problem in Eq. (10) involves a squared Frobenius norm and the nuclear norm. It is of the form

$$\underset{\mathbf{Z}}{\operatorname{argmin}} \frac{1}{2}\|\mathbf{B} - \mathbf{Z}\|_{\mathcal{F}}^2 + \eta\|\mathbf{Z}\|_* \qquad (11)$$

and can be solved by a soft-impute algorithm (in our case, $\eta = \theta\tau$). We rewrite the nuclear norm as

$$\|\mathbf{Z}\|_* := \min_{\mathbf{U},\mathbf{V}:\ \mathbf{Z}=\mathbf{U}\mathbf{V}} \frac{1}{2}\left(\|\mathbf{U}\|_\mathcal{F}^2 + \|\mathbf{V}\|_\mathcal{F}^2\right). \quad (12)$$

The solution to this problem is given by $\mathbf{Z} = \mathbf{U}\mathbf{D}_\eta\mathbf{V}$, where $\text{svd}(\mathbf{Z}) = \mathbf{U}\mathbf{D}\mathbf{V}$, $\mathbf{D} = \text{diag}(\sigma_1, ..., \sigma_r)$ and

$$\mathbf{D}_\eta = \big(\max(\sigma_1 - \eta, 0), ..., \max(\sigma_r - \eta, 0)\big). \quad (13)$$

The energy in Eq. (9) is convex, but — because of the TV regularizer — not differentiable. Nevertheless, the problem can be dualised with Legendre-Fenchel transform. The primal-dual form is then given by

$$\operatorname*{argmin}_{\mathbf{S}} \max_q \frac{1}{2\theta}\|\mathbf{S} - \bar{\mathbf{S}}\|_\mathcal{F}^2 + \frac{\lambda}{2}\|\mathbf{W} - \mathbf{R}\mathbf{S}\|_\mathcal{F}^2 +$$
$$\frac{\gamma}{2}\|\mathbf{S} - \mathbf{S}_{\text{prior}}\|_\mathcal{F}^2 + \sum_{f,i,p}\left(\mathbf{S}_{fi}(p)\nabla^* q_f^i(p) - \delta\left(q_f^i(p)\right)\right), \quad (14)$$

where $q$ is the dual variable that contains the 2-dimensional vectors $q_f^i(p)$ for each frame $f$, coordinate $i$ and pixel $p$. $\nabla^* = -\text{div}(\cdot)$ is the adjoint of the discrete gradient operator $\nabla$, and $\delta$ is the indicator of the unit ball. In the primal-dual algorithm used to solve the problem, firstly the differential $\mathbf{D}_q$ of the dual part is initialised. Next, the gradient w.r.t. $\mathbf{S}$ is computed and set to zero to obtain a temporal minimizer $\bar{\mathbf{S}}$. Next, $\mathbf{D}_q$ is updated. The algorithm alternates between finding $\bar{\mathbf{S}}$ and updating $\mathbf{D}_q$ until convergence. The gradient operator $\nabla_\mathbf{S}$ applied to the energy in Eq. (14) yields

$$(\lambda\mathbf{R}^\mathsf{T}\mathbf{R} + \gamma + \frac{1}{\theta})\mathbf{S} - (\lambda\mathbf{R}^\mathsf{T}\mathbf{W} + \frac{1}{\theta}\bar{\mathbf{S}} + \gamma\mathbf{S}_{\text{prior}} - \mathbf{D}_q). \quad (15)$$

The minimizer $\bar{\mathbf{S}}$ is obtained by imposing $\nabla_\mathbf{S}(\cdot) \stackrel{!}{=} 0$ as

$$\left(\lambda\mathbf{R}^\mathsf{T}\mathbf{R} + \gamma + \frac{1}{\theta}\mathbf{I}\right)^{-1}(\lambda\mathbf{R}^\mathsf{T}\mathbf{W} + \frac{1}{\theta}\bar{\mathbf{S}} + \gamma\mathbf{S}_{\text{prior}} - \mathbf{D}_q). \quad (16)$$

An overview of the entire algorithm is given in Alg. 1. Note that **STEP 1** and **STEP 2** are repeated until convergence.

### 3.2. Per Frame Shape Prior

In the case of an inhomogeneous shape prior, *i.e.*, a shape prior different for every frame, the data term reads

$$\mathbf{E}_{\text{data}} = \frac{\gamma}{2}\|\Gamma(\mathbf{S} - \mathbf{S}_{\text{prior}})\|_\mathcal{F}^2, \quad (17)$$

where $\Gamma$ is a diagonal matrix controlling the influence of the shape prior for individual frames. Following the same principles as in Sec. 3.1, we derive the minimizer $\bar{\mathbf{S}}$ of the primal-dual formulation of Eq. (9) as

$$\left(\lambda\mathbf{R}^\mathsf{T}\mathbf{R} + \gamma\Gamma^\mathsf{T}\Gamma + \frac{1}{\theta}\mathbf{I}\right)^{-1}(\lambda\mathbf{R}^\mathsf{T}\mathbf{W} + \frac{1}{\theta}\bar{\mathbf{S}} + \gamma\Gamma^\mathsf{T}\Gamma\mathbf{S}_{\text{prior}} - \mathbf{D}_q). \quad (18)$$

Entries of $\Gamma$ adjust the parameter $\gamma$ framewise. If $\Gamma$ contains zero and non-zero values, it can be interpreted as a binary shape prior indicator for every frame[1].

---
[1] in the case if $\Gamma$ contains only zeroes and ones, $\Gamma^\mathsf{T}\Gamma = \Gamma$.

---

**Algorithm 1** SPVA: Variational NRSfM with a Shape Prior
**Input:** measurements $\mathbf{W}$, $\mathbf{S}_{prior}$, parameters $\lambda, \gamma, \tau, \theta, \eta = \theta\tau$
**Output:** non-rigid shape $\mathbf{S}$, camera poses $\mathbf{R}$
1: **Initialisation:** $\mathbf{S}$ and $\mathbf{R}$ under rigidity assumption [46]
2: **STEP 1.** Fix $\mathbf{S}$, find an optimal $\mathbf{R}$ *framewise*:
3: $\text{svd}(\mathbf{W}\mathbf{S}(\mathbf{S}\mathbf{S}^\mathsf{T})^{-1}) = \mathbf{U}\Sigma\mathbf{V}^\mathsf{T}$
4: $\mathbf{R} = \mathbf{U}\mathbf{C}\mathbf{V}^\mathsf{T}$, where
   $\mathbf{C} = \text{diag}(1, 1, ..., 1, \text{sign}(\det(\mathbf{U}\mathbf{V}^\mathsf{T})))$
5: **STEP 2.** Fix $\mathbf{R}$; find an optimal $\mathbf{S}$:
6: **while** not converge **do**
7:   **Primal-Dual:** fix $\bar{\mathbf{S}}$; *find an intermediate* $\mathbf{S}$ *(Eq. (9))*
8:   **Initialisation:** $q_f^i(p) = \mathbf{0}$
9:   **while** not converge **do**
10:     $\mathbf{D}_q = \begin{pmatrix} \nabla^* q_1^1(1) & \cdots & \nabla^* q_1^1(N) \\ \vdots & \ddots & \vdots \\ \nabla^* q_F^3(1) & \cdots & \nabla^* q_F^3(N) \end{pmatrix}$
11:     $\mathbf{S} = (\lambda\mathbf{R}^\mathsf{T}\mathbf{R} + \gamma + \frac{1}{\theta}\mathbf{I})^{-1}$
12:     $(\lambda\mathbf{R}^\mathsf{T}\mathbf{W} + \frac{1}{\theta}\bar{\mathbf{S}} + \gamma\mathbf{S}_{\text{prior}} - \mathbf{D}_q)$
13:     **for** $f = 1, ..., F; i = 1, ..., 3; p = 1, ..., N$ **do**
14:       $q_f^i(p) = \frac{q_f^i(p) + \sigma\nabla\mathbf{S}_f^i(p)}{\max(1, \|q_f^i(p)\| + \sigma\nabla\mathbf{S}_f^i(p)\|)}$
15:   **end while**
16:   **Soft-Impute:** fix $\mathbf{S}$; *find an intermediate* $\bar{\mathbf{S}}$ *(Eq. (10))*
17:   $\text{svd}(P(\mathbf{S})) = \mathbf{U}\mathbf{D}\mathbf{V}^\mathsf{T}$, where $\mathbf{D} = \text{diag}(\sigma_1, ..., \sigma_r)$
18:   $\bar{\mathbf{S}} = \mathbf{U}\mathbf{D}_\eta\mathbf{V}^\mathsf{T}$, where
   $\mathbf{D}_\eta = \text{diag}\big(\max(\sigma_1 - \eta, 0), ..., \max(\sigma_r - \eta, 0)\big)$
19: **end while**

### 3.3. Per Pixel per Frame Shape Prior

Per pixel per frame shape prior is the most general form of the proposed constraint; integration of it is more challenging. Firstly, we obtain the matrices $\tilde{\mathbf{S}} \in \mathbb{R}^{3FN}$, $\tilde{\bar{\mathbf{S}}} \in \mathbb{R}^{3FN}$, $\tilde{\mathbf{S}}_{\text{prior}} \in \mathbb{R}^{3FN}$ and $\tilde{\mathbf{D}}_q \in \mathbb{R}^{3FN}$ from $\mathbf{S}$, $\bar{\mathbf{S}}$, $\mathbf{S}_{\text{prior}}$ and $\mathbf{D}_q$ respectively, by applying the permutation operator $P(\cdot)$ and stacking point coordinates of all frames into a vector (*e.g.*, $\tilde{\mathbf{S}} = \text{vec}(P(\mathbf{S}))$, and analogously for the remaining matrices). Similarly, we obtain matrix $\tilde{\mathbf{W}} \in \mathbb{R}^{2FN}$ as

$$\big(\underbrace{\mathbf{W}_{11}\mathbf{W}_{21}\cdots\mathbf{W}_{1N}\mathbf{W}_{2N}}_{\text{all points of frame 1}}\cdots\underbrace{\mathbf{W}_{(2F-1)1}\mathbf{W}_{(2F)1}\cdots}_{\text{all points of frame }F}\big)^\mathsf{T}. \quad (19)$$

Accordingly, the rotation matrix is adjusted. The resulting matrix $\tilde{\mathbf{R}} \in \mathbb{R}^{2FN \times 3FN}$ is a quasi-block diagonal. It contains $FN$ blocks of size $2 \times 3$, *i.e.*, $\tilde{\mathbf{R}} = \text{diag}\{\boxed{\mathbf{R}_1}\ldots\boxed{\mathbf{R}_F}\ldots\}$. We introduce a diagonal matrix $\tilde{\Gamma} \in \mathbb{R}^{3FN \times 3FN}$ containing weights per frame per point coordinate. After applying proximal splitting, $P(\cdot)$ and $\text{vec}(\cdot)$ operators, the minimisation problem in Eq. (9) alters to

$$\operatorname*{argmin}_{\tilde{\mathbf{S}}} \frac{\lambda}{2}\|\tilde{\mathbf{W}} - \tilde{\mathbf{R}}\tilde{\mathbf{S}}\|_\mathcal{F}^2 + \frac{1}{2\theta}\|\tilde{\mathbf{S}} - \tilde{\bar{\mathbf{S}}}\|_\mathcal{F}^2 +$$
$$\frac{\gamma}{2}\|\tilde{\Gamma}(\tilde{\mathbf{S}} - \tilde{\mathbf{S}}_{\text{prior}})\|_\mathcal{F}^2 + \sum_{f,i,p}\|\nabla\mathbf{S}_f^i(p)\|. \quad (20)$$

The gradient of the function in Eq. (20) reads

$$\nabla_{\tilde{\mathbf{S}}} = (\lambda \tilde{\mathbf{R}}^\top \tilde{\mathbf{R}} + \frac{1}{\theta}\mathbf{I}_{3FN} + \gamma \tilde{\Gamma}^\top \tilde{\Gamma})\tilde{\mathbf{S}} -$$

$$(\lambda \tilde{\mathbf{R}}^\top \tilde{\mathbf{W}} + \frac{1}{\theta}\bar{\tilde{\mathbf{S}}} + \gamma \tilde{\Gamma}^\top \tilde{\Gamma}\tilde{\mathbf{S}}_{\text{prior}} - \tilde{\mathbf{D}}_q) \stackrel{!}{=} 0. \quad (21)$$

Finally, the minimizer of Eq. (21) is obtained as

$$\tilde{\mathbf{S}} = (\underbrace{\lambda \tilde{\mathbf{R}}^\top \tilde{\mathbf{R}}}_{\text{block-diagonal}} + \underbrace{\frac{1}{\theta}\mathbf{I}_{3FN}}_{\text{diagonal}} + \underbrace{\gamma \tilde{\Gamma}^\top \tilde{\Gamma}}_{\text{diagonal}})^{-1}$$

$$(\lambda \tilde{\mathbf{R}}^\top \tilde{\mathbf{W}} + \frac{1}{\theta}\bar{\tilde{\mathbf{S}}} + \gamma \tilde{\Gamma}^\top \tilde{\Gamma}\tilde{\mathbf{S}}_{\text{prior}} - \tilde{\mathbf{D}}_q). \quad (22)$$

Note that the factor on the left side of Eq. (22) represents a block-diagonal matrix. Its inverse can be found by separately inverting $FN$ blocks of size $3 \times 3$. After $\tilde{\mathbf{S}}$ is computed, we obtain $\mathbf{S}$ by an inverse permutation.

## 4. Obtaining Shape Prior

In this section, we revise the method for occlusion tensor estimation, and formulate a criterion for a set of views to be suitable for the shape prior estimation; more details are placed in Appendix, parts A and B.

**Occlusion tensor estimation.** An occlusion tensor is a probabilistic space-time occlusion indicator. We refer to occlusion maps as slices of the occlusion tensor corresponding to individual frames. If occlusion tensor is available, it is possible to control a shape prior with the per pixel per frame granularity (see Sec. 3.3). Occlusion tensor is computed from $\mathbf{W}$ and a reference image. For every frame, a corresponding occlusion map equals to a Gaussian-weighted difference between a backprojection of the frame to the reference frame and the reference frame itself. Thus, the occlusion indicator triggers a higher response for areas which cannot be backprojected accurately due to occlusions, specularities, illumination inconstancy, large displacements, highly non-rigid deformations, or a combination of those. As a result, the occlusion tensor accounts for multiple reasons of inaccuracies in correspondences. A similar scheme was applied in [32, 40]. The complexity of the occlusion tensor estimation is $\mathcal{O}(F\mathbf{whJ}^2)$ with $\mathbf{w}, \mathbf{h}$ and $\mathbf{J}$ being width and height of a frame and a size of a square Gaussian kernel respectively. For few dozens of frames of common resolutions as they occur in NRSfM problems, the whole computation can be performed on a GPU in less than a second. Examples of occlusion maps are given in Figs. 1, 4.

**Total intensity criterion.** Given an occlusion tensor, we determine the set of frames suitable for the shape prior estimation using the accumulative *total intensity criterion*:

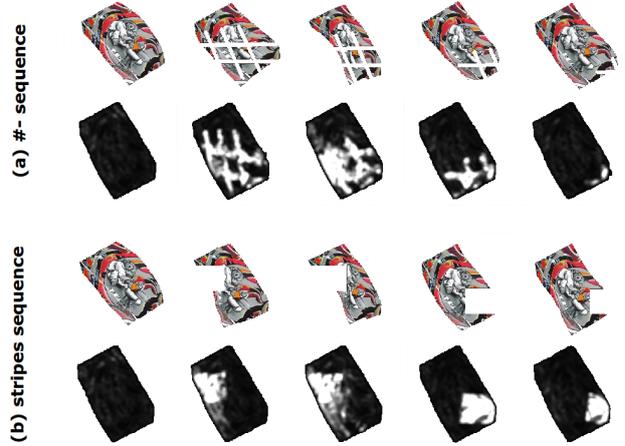

**Figure 1:** Exemplary frames from the modified flag sequences [16] with the computed occlusion maps: (a) #-sequence; (b) *stripes* sequence.

$$\sum_{f=1}^{F_{sp}} \left\| \int_\Omega du\, dv \right\|_2 \leq \epsilon. \quad (23)$$

In Eq. (23), $\Omega$ denotes an image domain of a single frame, $F_{sp}$ denotes length of the sequence suitable for shape prior estimation, and $\epsilon$ is a non-negative scalar value. In other words, as far as the frames are not significantly occluded (regardless in which image region occlusions happen), they can be used for the estimate. The obtained shape prior is rigidly aligned with an initialisation obtained with [46], since a different number of frames significantly affect initial alignment of the reconstructions. Therefore, we employ Procrustes analysis on 3D points corresponding to unoccluded image pixels ($15-20$ points are uniformly selected).

## 5. Experiments

The proposed approach is implemented in C++/CUDA C [25] for a heterogeneous platform with a multi-core CPU and a single GPU. We run experiments on a machine with Intel Xeon E5-1650 CPU, NVIDIA GK110 GPU and 32 GB RAM. While finding an optimal $\mathbf{R}$ (Eqs. (4)–(6)), the most computationally expensive operation is the product $\mathbf{SS}^\top$. This operation can be accomplished by six vector dot products and only $\mathbf{S}$ needs to be stored in memory. It is implemented as a dedicated GPU function, together with the computation of $\mathbf{D}_q$ (Alg. 1, rows 10, 13, 14). Compared to the C++ version, $12-15\mathrm{x}$ speedup is achieved. To compute dense correspondences, Multi-Frame Subspace Flow (MFSF) [16] is used, and to estimate a shape prior, we run [15] on several initial frames of the sequence as described in Sec. 4. If not available for a respective sequence, segmentations of the reference frames are computed with [34].

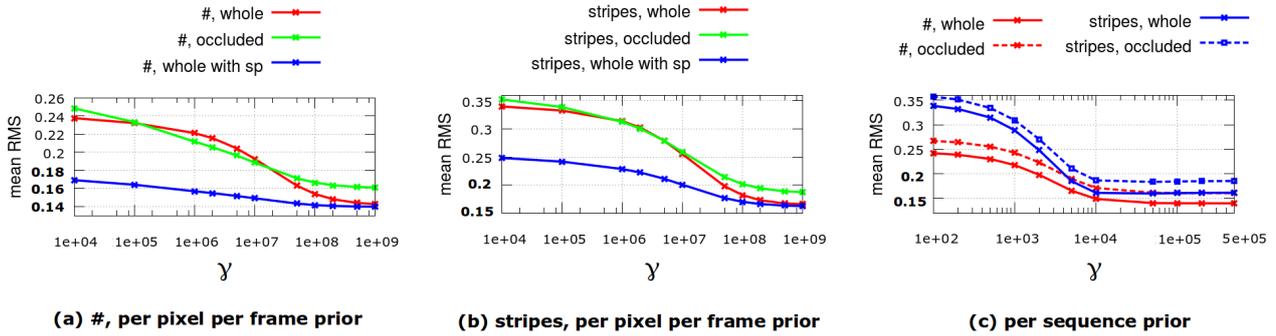

**Figure 2:** Results of the quantitative evaluation of the proposed method in the configuration MFSF[16] + SPVA: (a) per pixel per frame mode on the #-sequence; (b) per pixel per frame mode on the *stripes* sequence; (c) per sequence mode on both sequences. "whole": mean RMS is computed on all frames of the respective sequence, "occluded": mean RMS is computed only on the occluded frames; "whole with sp": the algorithm is initialised with the shape prior in the non-occluded frames. Bold font (mean RMS) highlights parameter values which outperform occlusion-aware MFOF[40] + VA[15].

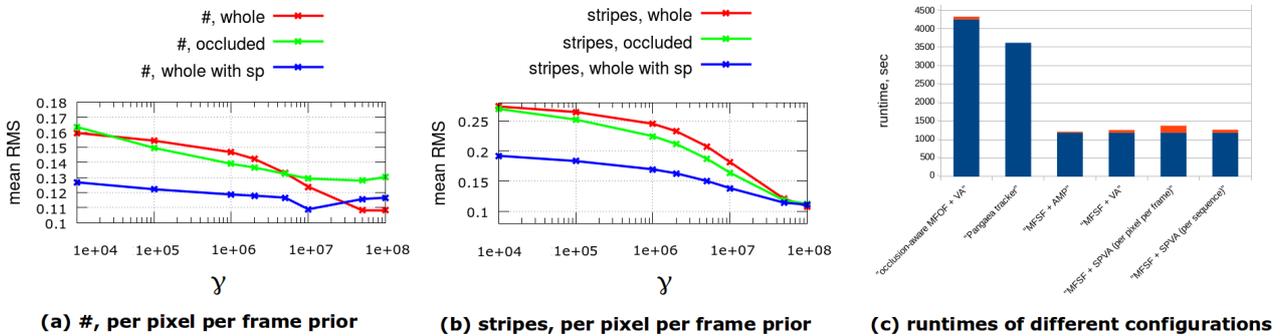

**Figure 3:** Results of the quantitative evaluation on the flag sequence with the dense segmentation mask. In (a) and (b), the notation is the same as in Fig. 2. Reconstructions obtained on the unoccluded ground truth optical flow are used as a reference for comparison; (c) runtimes of different pipeline configurations on the dense flag dataset (blue color marks correspondence computation, orange marks NRSfM, except for Pangaea which is a template-based method). The fastest configuration MFSF[16] + AMP[17] is only ca. 4% faster than the proposed configuration with SPVA which is the most accurate.

As our objective is to jointly evaluate correspondence establishment under severe occlusions and non-rigid reconstruction, we perform joint evaluation of different pipeline configurations. We compare occlusion-aware MFOF[40] + VA[15], MFSF[16] + AMP[2][17], MFSF + VA, MFSF + SPVA (the proposed method). For every configuration, we report mean Root-Mean-Square (RMS) error metric defined as $e_{3D} = \frac{1}{F}\sum_{f=1}^{F} \frac{\|\mathbf{S}_f^{ref} - \mathbf{S}_f\|_\mathcal{F}}{\|\mathbf{S}_f^{ref}\|_\mathcal{F}}$, where $\mathbf{S}_f^{ref}$ is a ground truth surface in 3D. Subsequently, we show results on real image sequences and compare results qualitatively.

**Evaluation Methodology.** For the joint evaluation, a dataset with a ground truth geometry and corresponding images is required. There is one dataset known to the authors which partially fulfils the requirements — the synthetic flag sequence initially introduced in [16]. This dataset originates from mocap measurements and contains images of a waving flag rendered by a virtual orthographic camera. The flag dataset was already used for evaluation of NRSfM [1, 3] and MFOF algorithms [16, 40], but not for a joint evaluation, to the best of our knowledge. To generate orthographic views, the mocap flag data was projected onto an image plane (with an angle of approx. $30°$ around the $x$ axis) and a texture was applied on it (here the texture does not reflect distortion effects associated with the view which is different from the frontal one). More details on the dataset can be found in [35]. Using the rendered images, we evaluate MFOF and NRSfM methods jointly.

First, we extend the flag dataset with several data structures. The ground truth surfaces contain 9622 points, whereas the rendered images are of the resolution $500 \times 500$. If the corresponding segmentation mask for the reference frame is applied, $8.2 \cdot 10^4$ points are fetched. To overcome this circumstance, we create a segmentation mask which fetches the required number of points as in the ground truth. Therefore, we project the ground truth surface corresponding to the reference frame onto the image plane and obtain a sparse segmentation mask. When applied to the dense

---
[2]AMP is an optimised extension of Metric Projections [27].

**W**, the sparse mask fetches 9622 points. To establish point correspondences between the ground truth and reconstructions, we apply non-rigid point set registration with correspondence priors [18]. This procedure needs to be preformed only once on a single ground truth surface and a single flag reconstruction with 9622 points, and the correspondence index table is used during computation of the mean RMS. Non-rigid registration does not alter any reconstruction which is evaluated for mean RMS.

Second, we introduce severe occlusions into the flag image sequence which go beyond those added for evaluation in [40] in terms of the occlusion duration and size of the occluded regions. We overlay two different patterns with the clean flag sequence — a grid (#) and stripes patterns. The resulting sequences contain 20 and 29 occluded frames respectively (see Fig. 1 for exemplary frames and the corresponding occlusion maps).

**Experiments on synthetic data.** We compare several framework configurations on the synthetic flag sequence and report mean RMS and runtimes. We also evaluate the influence of the shape prior term through varying the $\gamma$ parameter in several shape prior modes.

Results of the experiment are summarised in Fig. 2. Occlusion-aware MFOF+VA achieves the mean RMS error of 0.18(0.219) (in brackets, mean RMS only on occluded frames is reported) for the #-sequence and 0.195(0.209) on the *stripes*. MFSF+VA achieves 0.239(0.256) and 0.341(0.355) for the #- and *stripes* sequences respectively. MFSF+SPVA achieves 0.143(0.161) and 0.167(0.187) for the #- and *stripes* sequence respectively in the per pixel per frame mode, and 0.140(0.160) and 0.160(0.183) in the per sequence mode. At the same time, runtime of the MFSF+SPVA in the per frame mode is almost equal to the runtime of MFSF+VA — the difference is less than 1% — whereas the configuration MFSF+SPVA in the per pixel per frame mode takes only 3% more time.

The configuration with the fastest MFSF and the proposed SPVA achieves the lowest mean RMS; it is comparable in the runtime to the fastest configuration with AMP. We are 3.4 times faster than the second best NRSfM based configuration with the computationally expensive occlusion-aware MFOF+VA. As can be seen in Fig. 2, performance of SPVA depends on $\gamma$. As expected, mean RMS is the lowest for a particular finite $\gamma$ value and grows as $\gamma$ increases. The drop in the accuracy happens because the shape prior term becomes so dominant that even less probably occluded areas are regularised out. If $\gamma$ is infinite, all frames (all pixels with non-zero occlusion map) are set to the shape prior which leads to a suboptimal solution. In the per sequence mode (Fig. 2-(c)) $\gamma \in [0; 10^6]$, whereas in per pixel per frame mode $\gamma \in [0; 10^9]$ ($\gamma$ is split between all pixels weighted with the occlusion map values). Experi-ment shows that the transitions are gradual with the gradual changes of $\gamma$.

| algorithmic combination | mean RMS, # | mean RMS, *stripes* |
|---|---|---|
| o.a. MFOF [40]+VA [15] | 0.181 (0.219) | 0.195 (0.209) |
| Pangaea tracker [51] | **0.172 (0.191)** | **0.172 (0.191)** |
| MFSF [16]+AMP [17] | 0.297 (0.381) | 0.460 (0.523) |
| MFSF [16]+VA [15] | 0.239 (0.252) | 0.341 (0.355) |
| MFSF [16]+SPVA, p. pix. | **0.143 (0.161)** | **0.167 (0.189)** |
| MFSF [16]+SPVA, p. seq. | **0.140 (0.160)** | **0.160 (0.184)** |

**Table 1:** Mean RMS errors of different algorithmic combinations for the #- and *stripes* sequences.

Besides, we perform a comparison of our method with the recent template-based method of Yu *et al.* [51] — Pangaea tracker. SPVA can be classified as a hybrid method for monocular non-rigid reconstruction. Firstly, the assumption of a template-based technique — an exact 3D shape is known for at least a single frame — is not fulfilled in our case (a shape prior is not an exact reconstruction). Secondly, we obtain shape prior automatically, whereas in template-based methods [38, 28, 9, 45, 51], a template is assumed to be known in advance. Nevertheless, the comparison with such a method as [51] is valuable. Ultimately, research in the area of template-based reconstruction shifts in the direction of hybrid methods, *i.e.*, there is an endeavour to find a way to obtain a template automatically and under non-rigid deformations. The SPVA framework is perhaps the first attempt in this direction, and in this experiment, we demonstrate that a template-based method can work with a shape prior obtained with the proposed approach (see Sec. 4) and produce accurate results. Pangaea tracker achieves almost equal mean RMS of 0.172 (0.191) for both #- and *stripes* sequences. We discovered that this template-based method is stable against textureless occlusions, but an error may accumulate if occlusions are permanent and large. Still, Pangaea tracker achieves the second best result after the combination MFSF+SPVA and outperforms the more recent occlusion-aware MFOF+VA pipeline on the sparse flag sequence. Table 2 summarizes the lowest achieved mean RMS errors for all tested combinations.

The experiment with the varying $\gamma$ is also repeated on the flag sequence with the dense segmentation mask. Here, we obtain reference reconstructions for comparison on the ground truth optical flow available for the unoccluded views. In this manner, it is possible to see how good the proposed pipeline alleviates side effects associated with occlusions and how close the reconstructions reach the reference. Moreover, the TV term is enabled, since the measurements are dense. Results are summarised in Fig. 3. The mean RMS relative to the reference reconstruction follows the similar pattern as in the case of the comparison with the sparse ground truth. In Fig. 3-(c), runtimes for all tested pipeline configurations are summarised.

In both experiments, a relatively high mean RMS

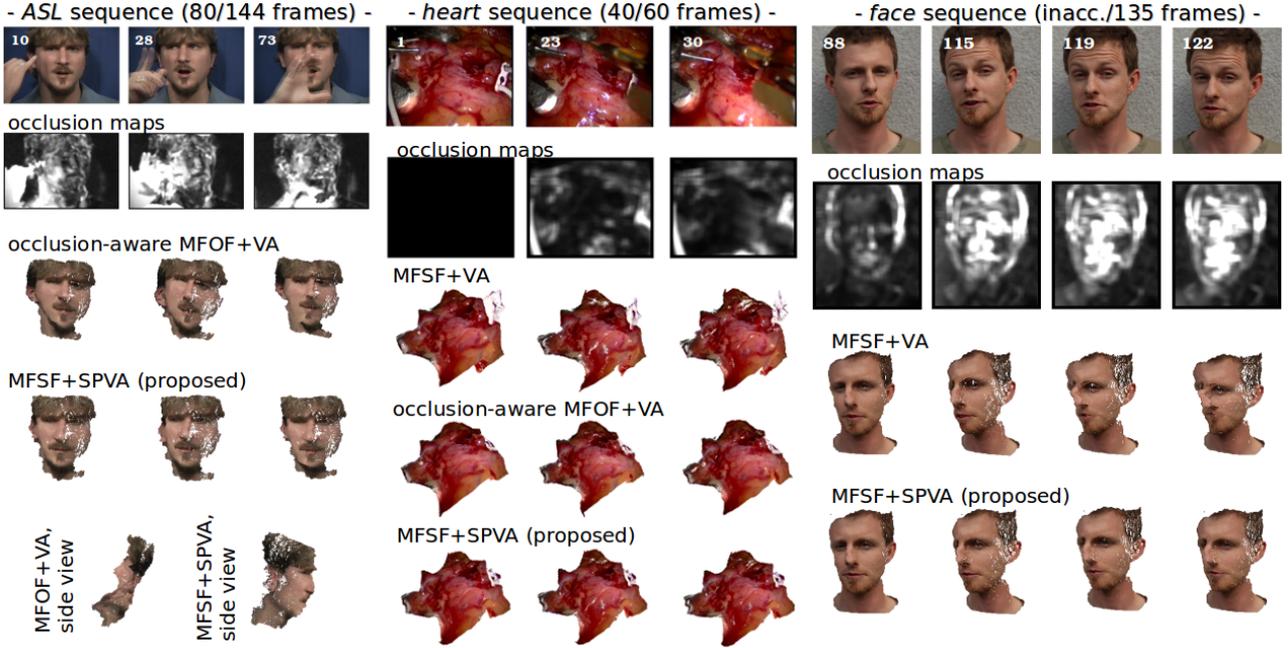

**Figure 4:** Qualitative results of the proposed SPVA framework and other pipeline combinations on several challenging real image sequences.

is explained by two effects. As above mentioned, the reference frame is not a frontal projection of the ground truth, and no frontal views are occurring in the image sequence. Moreover, the flag sequence exhibits rather large non-rigid deformations. All evaluated methods including the proposed approach perform best if deformations are moderate deviations from some mean shape.

**Experiments on real data.** We tested SPVA on several challenging real-world image sequences: *American Sign Language (ASL)* [10], *heart surgery* [39], and a new *face*. Results are visualised in Fig. 4. *ASL* sequence depicts a face with permanent occlusions due to hand gesticulation. Only sparse reconstructions were previously shown on it [19, 20]. On this sequence, occlusion-aware MFOF performs poorly and marks whole frames starting from frame 20 as occluded. Consequently, the combination MFOF[40]+VA fails to reflect realistic head shapes, and it is seen distinctly in the side view. The proposed approach, using the shape prior obtained on first 17 frames provides realistic reconstructions. The *heart* sequence is a recording of a heart bypass surgery. 40 out of 60 frames are significantly occluded by a robotic arm. For the first time, dense reconstructions on this sequence were shown in [40]. The proposed SPVA achieves similar appearance, but the runtime is 30% lower. The new *face* sequence depicts a speaking person. No external occlusions are happening, but MFSF produces noisy correspondences due to large head movements. Thus, MFSF+VA outputs reconstructions with a bent structure in the nose area, whereas the shape prior in SPVA suppresses unrealistic twisting. More details on the experiments with the real data can be found in Appendix C.

## 6. Conclusions

In this paper, we proposed the SPVA framework — a new approach for dense NRSfM which is able to handle severe occlusions. Thanks to the shape prior term, SPVA penalizes deviations from a meaningful prior shape. The highest supported granularity is per frame per pixel. The shape prior is automatically obtained on the fly from several non-occluded frames under non-rigidity using the total intensity criterion. The new approach does not require any predefined template or a deformation model. Along with that, we analysed relation to the template-based monocular reconstruction and came to the conclusion that SPVA can be considered as a hybrid method. A new evaluation methodology was introduced allowing to jointly evaluate correspondence computation and non-rigid reconstruction. Experiments showed that the proposed framework can efficiently handle scenarios with large permanent occlusions. The SPVA pipeline outperformed the baseline occlusion-aware MFOF+VA in terms of accuracy and runtime. A limitation of the proposed method lies in its pipeline nature — it can recover from inaccuracies in the pre-processing steps only up to a certain degree. Future work considers an extension to handle perspective distortions and a search for an optimal operation scheme for interactive processing.

# Appendix

This part provides insights into SPVA going beyond the scope of the main matter. In Secs. A and B, more details on occlusion tensor estimation and Total Intensity (TI) criterion are provided. Sec. C contains a more detailed description of the experimental results on the *human face*, *heart surgery*, and *ASL* sequences. We show that on the ASL sequence, the correspondence correction [40] in combination with the base scheme [15] fails, and analyse the reasons for failure. Moreover, algorithm parameters and runtimes for every sequence are listed.

## A. Obtaining Occlusion Tensor

Consider dense flow fields or displacements of pixel visible in the reference frame throughout the whole image sequence computed by MFSF [16]:

$$\mathbf{u}(x;\mathbf{n}) = \begin{bmatrix} u(x,\mathbf{n}) \\ v(x,\mathbf{n}) \end{bmatrix} : \Omega \times \{1, ..., F\} \to \mathbb{R}^2, \quad (24)$$

where $\Omega \in \mathbb{R}^2$, $\mathbf{n}$ is a frame index, $\mathbf{u}(x, \cdot)$ denotes a 2D displacement of a point $x$ through the image sequence ($u(x, \mathbf{n})$ and $v(x, \mathbf{n})$ denote displacements in $u$ and $v$ directions respectively). Let $\mathbf{r}$ be the index of the reference frame and $\mathbf{I}(x, \mathbf{r})$ the reference frame. Occlusion maps $\mathbf{E}(x, \mathbf{n}) : \Omega \times \{2, ..., F\} \to \mathbb{R}$ can be obtained from the dense correspondences and the reference frame according to the algorithm summarised in Alg. 2. Firstly, a backprojection $\mathbf{B}(\mathbf{n}, \mathbf{r})$ of every frame $\mathbf{I}(x, \mathbf{n})$ to the reference frame is performed. In this step, reverse point displacements are applied to every frame $\in \{2, \ldots, F\}$ with an optional interpolation for missing parts. Secondly, image differences between the warped images and the reference image are computed. Therefore, we take $\ell_2$ norms of sums of channel-wise differences (RGB) for every pixel convolved with the Gaussian kernel $G_{k \times k}$ of an odd width $k$ (see Alg. 2, rows 3–7). The result is post-processed (normalised and discretised) so that the estimated values lie in the interval $[0; 255]$. The resulting image series represents an occlusion tensor with per-pixel occlusion probabilities for every frame. If required, occlusion maps can be binarised. The entire algorithm exhibits data parallelism (both on the frame and pixel levels) and is well suitable for implementation on a GPU.

## B. Total Intensity Criterion

In Fig. 5, results of the algorithm operating based on the TI criterion are given. TI accumulates pixel intensities for all frames until the frame $f$[3]. As can be seen in the comparison of the images and corresponding TI plots, sudden increases in TI happen if an occlusion begins. Low occlusion

[3]in this sense, the TI criterion is similar to a cumulative distribution function.

**Algorithm 2** Estimation of an Occlusion Tensor $\mathbf{E}(x)$

**Input:** dense flow fields $\mathbf{u}(x; \mathbf{n})$, a reference frame $\mathbf{I}(x, \mathbf{r})$, Gaussian kernel $G_{k \times k}$
**Output:** occlusion maps $\mathbf{E}(x, \mathbf{n})$
1: **for** every frame $\mathbf{n} \in \{2, \ldots, F\}$ **do**
2:    $\mathbf{w}(\mathbf{n}, \mathbf{r}) = \mathbf{I}(x, \mathbf{n}) - \mathbf{u}(x; \mathbf{n})$ (backprojection to $\mathbf{I}(x, \mathbf{r})$)
3:    image difference $\mathbf{B}(x) = \mathbf{w}(\mathbf{n}, r) - \mathbf{I}(x, \mathbf{r}) \hat{=}$
4:    **for** every pixel $x$ **do**
5:       $\mathbf{B}(x) = \left\| (x_r^\mathbf{w} - x_r^\mathbf{I})^2 + (x_g^\mathbf{w} - x_g^\mathbf{I})^2 + (x_b^\mathbf{w} - x_b^\mathbf{I})^2 \right\|_2$
6:    **end for**
7:    $\mathbf{E}(x, \mathbf{n}) = \mathbf{B}(x) * G$ (convolution)
8:    postprocess $\mathbf{E}(x)$ (*e.g.*, binarise if required)
9: **end for**

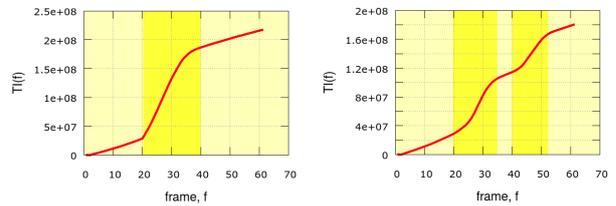

**(a)** TI plots for #-sequence (left) and *stripes* sequence (right). Deep yellow and light yellow colours mark occluded and unoccluded frames respectively. Note the difference in the function slopes for the occluded and unoccluded regions as well as the correlation between the slope changes and the beginning of occlusions in (b).

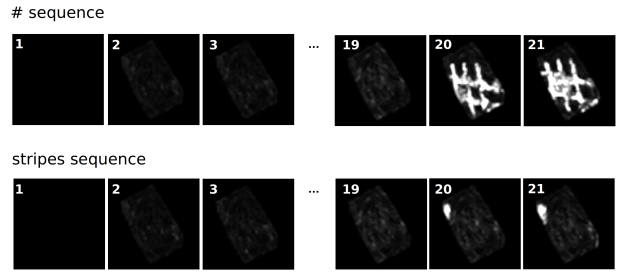

**(b)** Occlusion maps at the beginning of the sequences and when occlusions start.

**Figure 5:** Plots of TI function of the number of frames for #- and *stripes* sequences (a), excerpts from the sequences when occlusions begin (b).

probabilities cause gradual increases of TI. When comparing the frames with occlusion transitions and the responses of the TI indicators, the correlation is clearly seen.

The sensitivity threshold $\epsilon$ in Eq. (23) depends on the size and duration of an occlusion. Suppose occlusions do not happen and the sequence is infinitely long. In this case, $\epsilon$ will still be reached after a finite number of frames. This property is still desirable because the length of the sequence for the shape prior estimation should be narrowed down. TI can be augmented with or in some cases replaced by a

differential TI criterion such as

$$\frac{\text{TI}(F_{\text{sp}} + 1) - \text{TI}(F_{\text{sp}} - 1)}{2} \leq \epsilon', \quad (25)$$

where $\epsilon'$ is a threshold on the derivative value, and $F_{\text{sp}}$ is the last frame suitable for the shape prior estimation.

## C. Experiments on Real Sequences

In this section, more details on the experiments with real image data are given.

**The heart surgery sequence.** The heart sequence originates from [39] and shows a patient's heart during bypass surgery naturally non-rigidly deforming. The sequence contains 60 frames; at frame 20, a robotic arm enters the scene and occludes large regions of the scene over multiple frames. The shape prior is estimated on 18 initial frames. We use an average occlusion map for every frame since some of the regions disappear or are occluded in most of the frames. The results of the experiment are shown in Fig. 6. Due to a noisy initialisation under rigidity assumption, the combination MFSF[16]+VA[15] produces reconstructions with severe inaccuracies and discontinuities (Fig. 6-(c)). Occlusion-aware MFOF[40]+VA[15] generate visually consistent and smooth reconstructions but we notice that natural non-rigid deformations of the heart are attenuated (due to oversmoothing). MFSF[16]+SPVA produce visually consistent and accurate reconstructions which better reflect heart contractions.

**The human face sequence.** The new human face sequence depicting a speaking person was recorded with a monocular RGB camera. It contains 135 frames. Due to large displacements and deformations, MFSF [16] produces inaccurate correspondences especially around the frame 120. The direct method relying on data term [15] corrupts the structure and does not preserve point topology.

Using a per pixel per frame shape prior, we are able to preserve the point topology in the corrupted regions while relying on the data term where correspondences are accurate. Fig. 8 illustrates several problematic frames, corresponding occlusion maps and reconstructions with combinations MFSF[16]+VA[15] and MFOF[40]+SPVA. The experiment shows that even in the absence of occlusions a method for correspondence computation may perform poorly. In this case, reconstructions may exhibit such artefacts as broken point topology leading to corrupted reconstructions. Especially if it is not possible to recompute correspondences or available methods for obtaining correspondences do not improve the situation, our framework can be advantageous. For non-occluded regions (or regions with accurate correspondences), our approach strongly relies on the data term, whereas otherwise it strongly relies on the regularisation and shape prior terms.

**The ASL sequence.** The American Sign Language (ASL) sequence F5_10_A_H17 is taken from [10]. It shows a communicating person and contains severe occlusions due to hand gesticulation. Out of 114 frames, 80 frames have occlusions. To compute a shape prior, first 12 frames are used.

The processing results are shown in Fig. 9. The combination MFSF[16]+VA[15] suffers from the complexity of the seemingly simple scene (Fig. 9-(c),(e)). In this experiment, the occlusion-aware MFOF with correspondence correction [40] could not improve the reconstruction accuracy. The reason is a mixture of a suboptimal reference view, occlusions due to significant head rotations and severe external occlusions. Our approach paired with MFSF is the only one which achieves a meaningful reconstruction on the F5_10_A_H17 sequence, perhaps for the first time in the dense case (Fig. 9-(d), (f)). In Fig. 7, additional results are shown. In comparison to the reconstructions achieved by the combination MFSF[16]+VA[15] (see Fig. 7-(a)), the combination MFOF[40]+VA[15] (see Fig. 7-(b)) deteriorates the results. MFOF [40] can compensate for occlusions of small durations. If occlusions are large and permanent, the built-in occlusion indicator of the method may fail. As a side effect, the measurements are often oversmoothed in these cases. Another reason is a high default sensitivity of the occlusion indicator. Note that we have not tuned parameters of the occlusion indicators because they are supposed to be universal. In this example, however, large regions of the scene are spuriously detected as occlusions and the correction step relies on erroneous data.

Table 2 contains a summary of the performed experiments including parameters of the sequences, types of the applied shape priors and runtimes for all combinations and sequences. Parameters of the proposed SPVA approach are summarised in Table 3. In all experiments, $\sigma$ was set to $1.0$, and per frame per pixel shape prior was used.

**Discussion.** In all experiments, the proposed method performed more accurately on uncorrected correspondences as the base scheme, with minor added runtime. In some cases, the proposed approach can even produce more realistic dynamic reconstructions, as in the case of the *heart surgery* and *ASL* sequences. An advantage compared to Taetz *et al.* [40] is that the correction of inaccuracies is not restricted to a predefined procedure based on Bayesian inference. Different methods can be used to generate occlusion tensor, also integrating prior knowledge about a scene. At the same time, the proposed pipeline is faster and more suitable for online operation in scenarios with severe occlusions. Moreover, the proposed scheme can enhance the reconstruction accuracy when the occlusion-aware MFOF [40] fails to correct correspondences or correspondences cannot be computed anew. A limitation of the new approach is the dependency on the shape prior. If it is cor-rupted, the overall accuracy deteriorates. If an operational setting is known in advance and repeated over and over, a statistical shape prior can be invoked, instead of a shape prior obtained on the fly. Additionally, a faster frame-to-frame method for dense correspondences could be used in this case, though fast and accurate correspondence computation is still an open question. Occlusion detection still requires a trajectory-based approach operating on multiple frames, but once correspondences are established, an occlusion tensor can be efficiently computed in parallel.

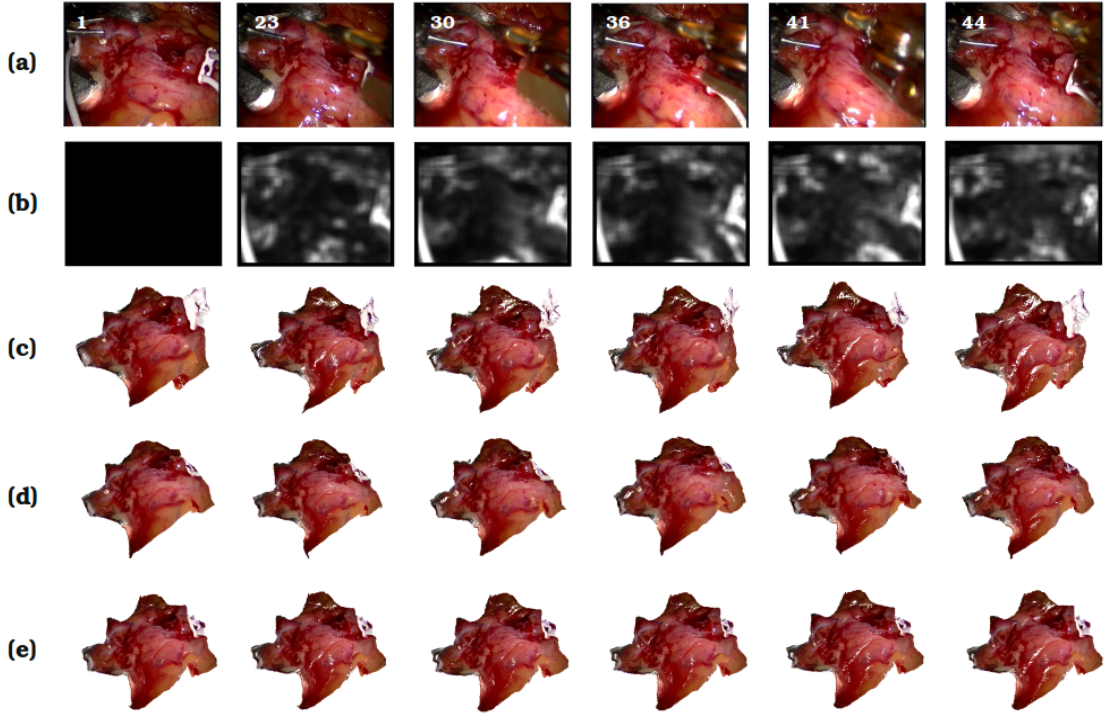

**Figure 6:** Experimental results on the *heart surgery* sequence [39]: (a) the reference frame 1 and several occluded frames; (b) per-pixel occlusion maps for the frames shown above; (c) reconstructions with MFSF[16]+VA[15]; (d): reconstructions with the occlusion-aware MFOF[40]+VA[15]; (e) reconstructions with MFSF[16]+SPVA (per frame).

| configuration | heart surgery [39] $360 \times 288$, 50 fr. | human face (new) $241 \times 285$, 136 fr. | ASL F5_10_A_H17 [10] $720 \times 480$, 114 fr. |
|---|---|---|---|
| MFSF [16] + VA [15] | 481.0 + 119.3 | 728.9 + 35.7 | 3114.0 + 400.0 |
| MFSF [16] + AMP [17] | 481.0 + 20.4 | 728.9 + 26.4 | 3114.0 + 98.0 |
| occlusion-aware MFOF [40] + VA [15] | 1592.8 + 119.2 | 2693.6 + 35.7 | 11995.3 + 300.5 |
| MFSF [16] + SPVA | 481.0 + 846.2 | 728.9 + 122.9 | 3114.0 + 1011.0 |

**Table 2:** Runtimes of different algorithm combinations for the sequences involved in the experiments, in seconds.

| sequence / param. | $\lambda$ | $\theta$ | $\tau$ | $\gamma$ |
|---|---|---|---|---|
| *heart surgery* [39] | $10^4$ | $10^{-5}$ | $10^4$ | $5 \cdot 10^4$ |
| *human face* | $5 \cdot 10^3$ | $10^{-5}$ | $5 \cdot 10^3$ | $10^3$ |
| *ASL* [10] | $5 \cdot 10^4$ | $10^{-5}$ | $4.2 \cdot 10^3$ | $10^5$ |

**Table 3:** Parameters of the proposed approach for different sequences.

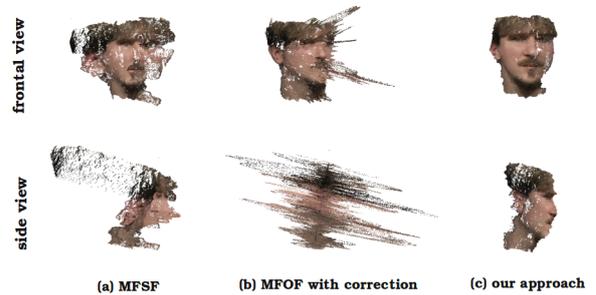

**Figure 7:** Results on the *ASL* sequence with correspondence correction; in this example, method by Taetz *et al.* [40] does not improve reconstruction accuracy; (a) reconstruction example of MFSF[16]+VA[15]; (b) reconstruction example of MFOF[40]+VA[15]; (c) reconstruction example of MFSF[16]+SPVA (the proposed approach).

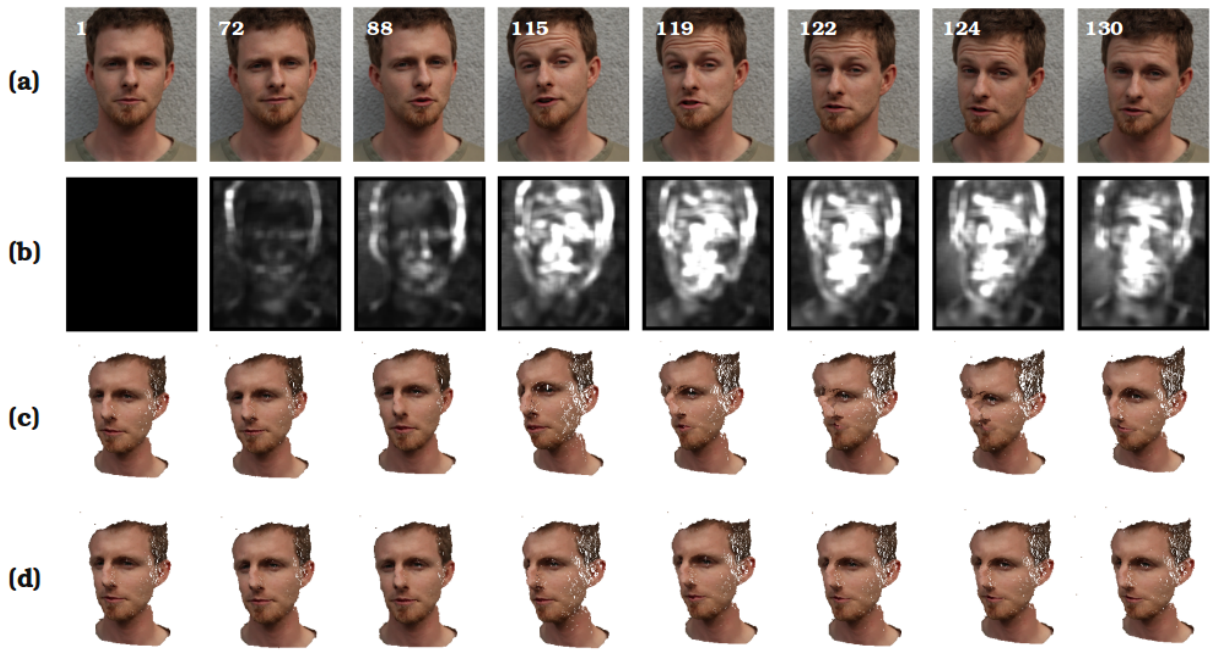

**Figure 8:** Experimental results on the *human face* sequence: (a) several frames of the sequence; (b) corresponding occlusion maps; (c) reconstructions with MFSF[16]+VA[15]; (d) reconstructions with MFSF[16] + SPVA (per frame per pixel).

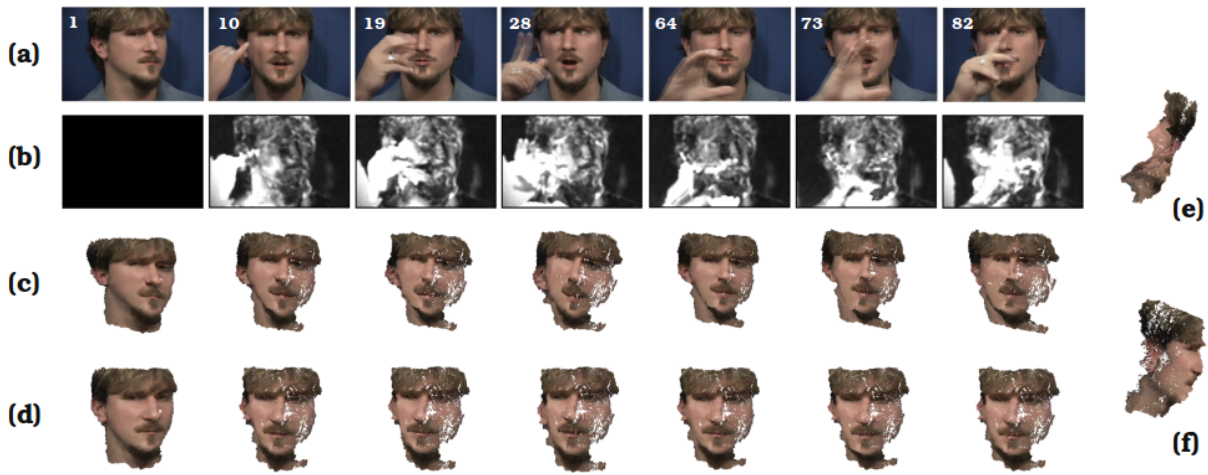

**Figure 9:** Experimental results for the *ASL* sequence [10]: (a) reference frame 1 and several frames with occlusions; (b) corresponding occlusion maps; (c) reconstructions with MFSF[16]+VA[15]; (d) reconstructions with MFSF[16]+SPVA (per frame per pixel); (e) side view of frame 28 for the configuration used in (c); (f) side view of frame 28 for the configuration used in (d).